\documentclass[letterpaper,10pt,journal,twoside]{IEEEtran}
\usepackage{amsmath,amsfonts}
\usepackage{amssymb}
\usepackage{amsthm}
\newtheorem{lemma}{Lemma}
\newtheorem{remark}{Remark}
\usepackage{algorithmic}
\usepackage{algorithm}
\usepackage{array}
\usepackage[caption=false,font=normalsize,labelfont=sf,textfont=sf]{subfig}
\usepackage{textcomp}
\usepackage{stfloats}
\usepackage{url}
\usepackage{verbatim}
\usepackage{graphicx}
\usepackage{cite}
\usepackage{units}
\usepackage{pgfplots}
\pgfplotsset{compat=1.18}
\usepgfplotslibrary{groupplots}
\usetikzlibrary{positioning} 
\usepackage{tikz}
\usetikzlibrary{calc,positioning,arrows.meta}
\usepackage{pgfplotstable}

\usepackage{tikz}
\usepackage{pgfplots}
\usepackage{pgfmath}

\usepackage{pgfplotstable}
\usepackage{circuitikz}
\usetikzlibrary{calc, bending, external, plotmarks, shapes, pgfplots.statistics, pgfplots.colorbrewer, positioning, arrows.meta, matrix, shapes.geometric}
\usepgfplotslibrary{fillbetween}
\pgfplotsset{scaled y ticks=false}
\pgfplotsset{compat=1.16}

\definecolor{matlab1}{rgb}{0.00000,0.44700,0.74100}
\definecolor{matlab2}{rgb}{0.85000,0.32500,0.09800}
\definecolor{matlab3}{rgb}{0.92900,0.69400,0.12500}
\definecolor{matlab4}{rgb}{0.49400,0.18400,0.55600}
\definecolor{matlab5}{rgb}{0.4660, 0.6740, 0.1880}
\definecolor{matlab6}{rgb}{0.3010, 0.7450, 0.9330}
\definecolor{matlab7}{rgb}{0.6350, 0.0780, 0.1840}
\definecolor{matlab8}{rgb}{0.8, 0.8, 0}
\definecolor{matlab9}{rgb}{0.6, 0.6, 0.6}

\begin{document}
\title{Mass Scaling Laws for Energy-Optimal Horizontal Multirotor Flight}

\author{Ayush Patnaik$^{1}$, Nicolas Michel$^{1}$, Leonard Bauersfeld$^{2}$, Davide Scaramuzza$^{2}$, and Xinfan Lin$^{1}$%
\thanks{Manuscript received: Feb 6, 2026; Revised: Sep 17, 2026; Accepted: Oct 18, 2026. This paper was recommended for publication by Editor Hyungpil Moon upon evaluation of the Associate Editor and Reviewers' comments \textit{(Corresponding author: Xinfan Lin).}}
\thanks{$^{1}$A. Patnaik, N. Michel, and X. Lin are with the Department of Mechanical and Aerospace Engineering, University of California, Davis, CA 95616 USA. {\tt\footnotesize ayupat@ucdavis.edu, lxflin@ucdavis.edu}}%
\thanks{$^{2}$L. Bauersfeld and D. Scaramuzza are with the Robotics and Perception Group, Department of Informatics, University of Zurich, 8050 Zurich, Switzerland. Their work was supported by the European Union's Horizon Europe Research and Innovation Programme under grant agreement No. 101120732 (AUTOASSESS).}%
\thanks{Digital Object Identifier (DOI): https://doi.org/10.1109/LRA.2026.3751905}

}

\markboth{IEEE Robotics and Automation Letters. Preprint Version. Accepted October, 2026}
{Patnaik \MakeLowercase{\textit{et al.}}: Mass Scaling Laws for Energy-Optimal Multirotor Flight}


\maketitle

\begin{abstract}
Multicopters are a type of aerial robot seeing extensive real-world applications
but with range and endurance as critical bottlenecks.
Optimization of its energy efficiency through 
energy-aware flight planning (e.g. cruise speed-scheduling)
has hence been an important topic.
The mass of the multicopter is an important factor to consider, 
as it has a significant impact on efficiency, and many common  
applications are subject to frequent mass variation due to payload change. 
This work derives closed-form scaling relationships between the energy-optimal steady horizontal flight behavior of a multicopter and its vehicle mass, and validates them experimentally on two multirotor platforms.
Specifically, for mass-only variation, i.e., 
when payload changes the vehicle mass while the airframe geometry, aerodynamic characteristics, and rotor configuration remain unchanged, 
we show through a combination of numerical simulation, theoretical derivation, and experimental validation that
(1) the energy-optimal velocity is proportional to the square root of vehicle mass; (2) the energy-optimal pitch angle is constant with respect to mass; and (3) the optimal energy per meter is proportional to mass.
The analysis shows that these three relationships arise from a common mass-similarity property of the steady-state flight 
and rotor-aerodynamic equations.
Furthermore, the obtained scaling relationships enable simple yet effective 
cruise-speed planning to optimize energy efficiency and extend the range and endurance of the multicopter.

\end{abstract}


\section{Introduction} 
\label{sec:Intro}

\IEEEPARstart{M}{ulticopters} have been emerging as a versatile aerial robot with wide applications, ranging from aerial photography and inspections, to intelligence, surveillance, and reconnaissance (ISR) missions, to package delivery among many others. 
However, their widespread adoption faces a major challenge in range and endurance, as their flight time, typically around 30 minutes, poses significant constraints on their capability and performance \cite{scaramuzza2022}. 

Prediction and optimization of multicopter energy consumption and performance have been actively studied in aerial robotics. Extensive work has focused on modeling energy dynamics and estimating flight range and endurance
\cite{scaramuzza2022, KIRSCHSTEIN2020102209, MURRAY2020368, POIKONEN2020104802, Stolaroff2018, FIGLIOZZI2017251, Li2024}.
While many early models focus on hovering or quasi-hovering conditions, recent studies consider the coupled rotor aerodynamics, propulsion, and rigid-body dynamics during forward flight
\cite{michel2022performance, scaramuzza2022}.
These aerodynamic effects make energy consumption strongly dependent on horizontal cruising velocity, with competing effects such as rotor inflow and body drag giving rise to an energy-optimal velocity.
\begin{figure}[t]
	\centering
    \scalebox{0.9}{\input{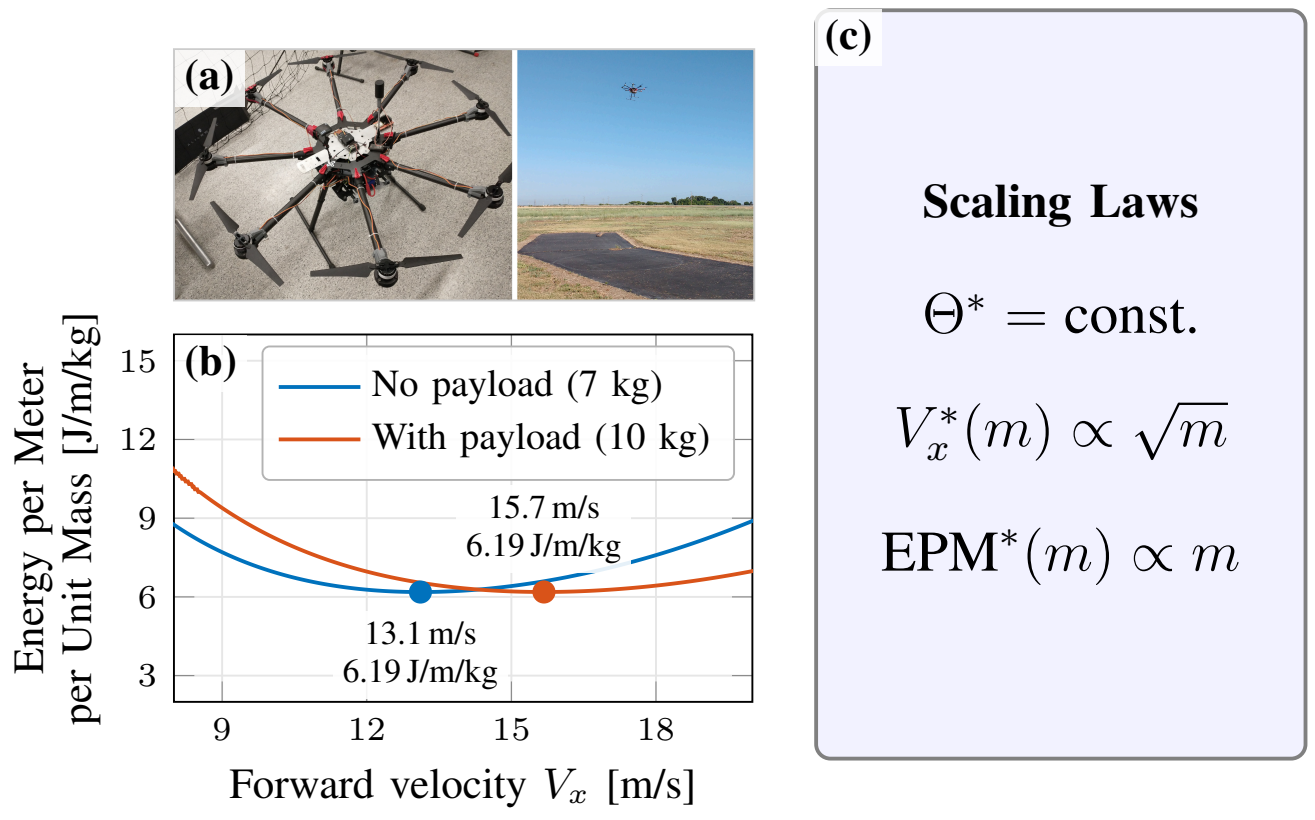}}
    \caption{Energy-optimal cruise behavior of a multicopter under payload-induced mass variation.
(a) Experimental octorotor platform used for outdoor flight tests, shown on the ground (left) and during steady horizontal flight (right).
(b) Theoretical energy consumption per unit distance normalized by mass (EPM/m) versus forward velocity for two vehicle mass configurations, demonstrating collapse of the mass-normalized energy efficiency at the energy-optimal operating point (marked).
(c) Scaling laws governing energy-optimal steady horizontal flight under mass variation for a multicopter: the optimal pitch angle is invariant to mass 
($\Theta^* = const.$), the optimal cruise velocity scales with the square root of mass ($V^*_x \propto \sqrt{m}$), and the minimum energy consumption per distance scales linearly with mass ($EPM^* \propto m$).}
	   \label{fig:Overall}
\end{figure}
In \cite{di2015energy}, experimental measurements are performed to quantify the difference in energy consumption caused by horizontal velocity, 
while several model-based studies, including \cite{michel2022performance, scaramuzza2022, DUKKANCI2021102985} among others, investigate the relationship using 
simulation. 
These works reveal the trade-off of various factors affecting the energy consumption rate, 
such as inflow aerodynamic effects and body drags, 
and the existence of an optimal velocity that maximizes the efficiency.  
Therefore, 
cruise-speed selection 
has become an important topic in energy-optimal flight planning and control 
\cite{scaramuzza2022,Zeng2019,di2015energy, wu2020inflight, Michel2023asme}.

One remaining topic of significant theoretical and practical interest is how energy efficiency and energy-optimal flight behavior change with vehicle mass.
In many multicopter applications, e.g. drone delivery and air mobility, vehicle mass is subject to frequent and substantial variations, making its impact on energy efficiency important.
Several existing works \cite{DUKKANCI2021102985,Hwang11092221,TAMKE2023,Patnaik2025} have investigated how the energy consumption versus velocity curve changes with mass through simulations and experiments.
Beyond the expected increase in energy consumption, these studies show that the energy-optimal velocity also increases with mass when other vehicle properties, including aerodynamics and geometry, remain unchanged.
The form of this dependence, however, has not been established.
In \cite{Hwang11092221}, it appears as a qualitative trend in a numerical endurance study as the payload-to-vehicle weight ratio is varied, with no explicit relationship stated.
In \cite{DUKKANCI2021102985}, the optimal speed is obtained numerically from an energy model calibrated to a reference vehicle, in which vehicle weight enters only the induced-power term while rotor speed, and hence blade profile power, is held fixed.
Neither work identifies the invariance of the energy-optimal pitch angle, from which the velocity scaling follows, or addresses the scaling of the minimum energy per meter.
Existing analytical models have also characterized the relationship among energy consumption, velocity, and vehicle mass.
A commonly used metric is Energy Per Meter (EPM), i.e., the energy consumed per unit distance.
For example, \cite{Dorling201770} extends a hovering-based energy model to horizontal flight to study the metric, 
while \cite{Dandrea2014} derives velocity-dependent power consumption for forward flight and obtains energy per distance by dividing power by velocity.
However, the former neglects forward-flight aerodynamic effects, while the latter simplifies them through a prescribed lift-to-drag relationship.
In summary, the quantitative relationships between vehicle mass and energy-optimal flight conditions remain to be established.

The contribution of this work is an in-depth theoretical, numerical, and experimental study of the relationships among energy efficiency, cruise velocity, and vehicle mass, as summarized in Fig.~\ref{fig:Overall}.
Building on established multicopter energy dynamics \cite{michel2019multiphysical, michel2022performance,scaramuzza2022},
we combine a propeller model based on blade-element/momentum theory (BEMT) with steady-state force balance from the rigid-body dynamics to derive explicit mass-scaling relationships for the energy-optimal cruise condition.
Specifically, for mass-only variation of a fixed vehicle configuration, i.e., when payload changes the vehicle mass while the airframe geometry, aerodynamic characteristics, and rotor configuration remain unchanged, we show
that
(1) the energy-optimal velocity is proportional to the square root of the mass;
(2) the energy-optimal pitch angle remains unchanged with mass;
and (3) the minimum energy consumption per unit distance, or EPM, is proportional to the vehicle mass.
These relationships are derived theoretically and validated experimentally on two multirotor platforms under payload-induced mass variations.

\section{Multicopter Energy Dynamics Modeling}

The theoretical and simulation studies in this work are based on a physics-based multicopter model that captures key coupled dynamics governing energy consumption, with a particular focus on rotor aerodynamics and rigid-body dynamics.
Strictly speaking, the analysis targets the mechanical and aerodynamic efficiency, which constitute the dominant contributors to energy consumption and are essential for characterizing energy-optimal cruise behavior. Other system components, including the motor, battery, and electronic speed controller (ESC), are not considered explicitly in the closed-form derivations, as their efficiencies are relatively constant over the operating regime of interest and have a comparatively smaller impact on the mass-scaling behavior. The influence of these components is discussed separately in Sec.~\ref{OCE}.

\subsection{Rigid Body Kinetics}
\begin{figure}[t]
	\centering
		\includegraphics[width=0.35\textwidth]{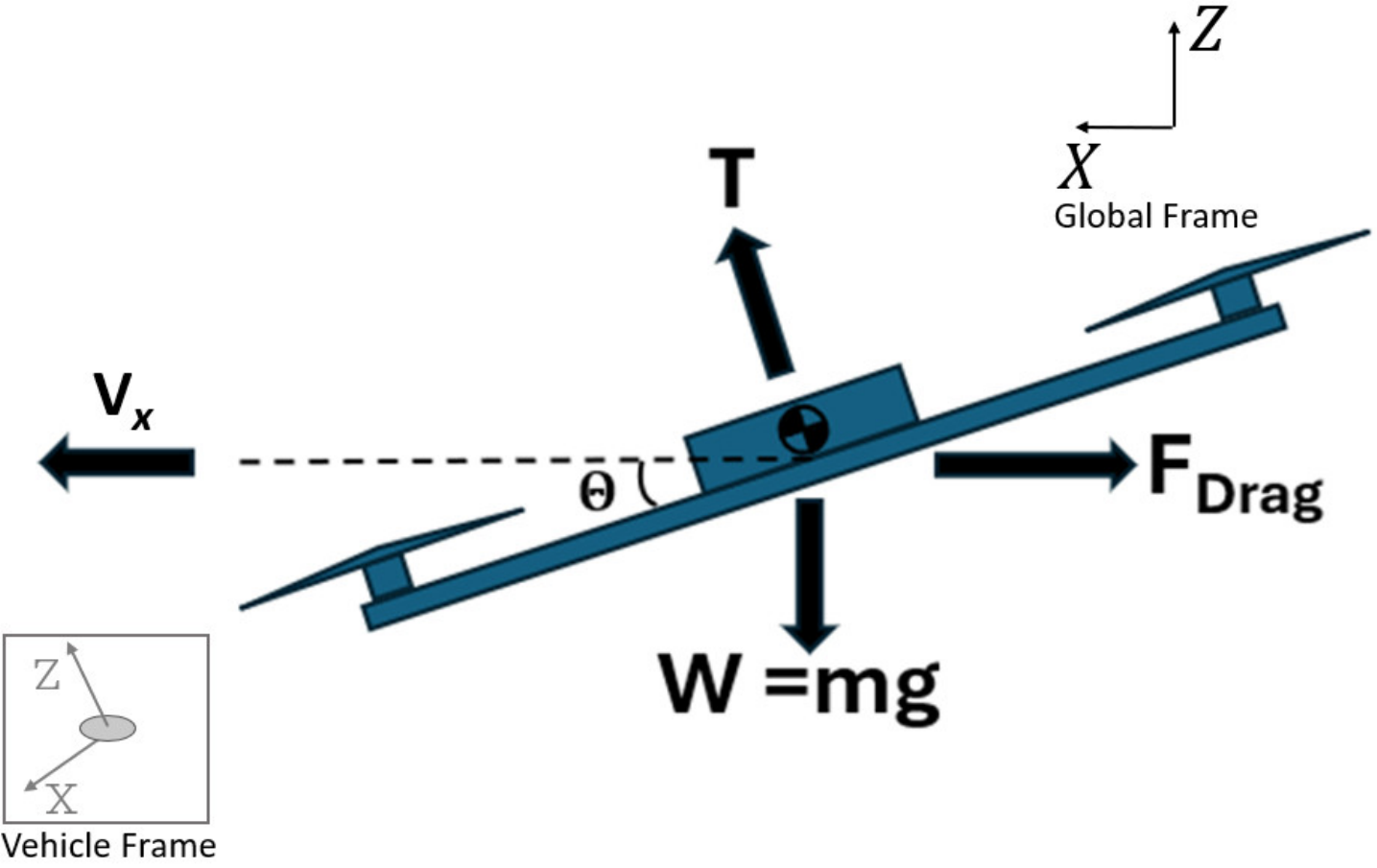}
	\caption{Multicopter Free Body Diagram }
	\label{fig: OctoFBD}
\end{figure}

The motion of the multicopter is modeled using the rigid body dynamics according to the free body diagram in Fig. \ref{fig: OctoFBD}. 
We consider 2D dynamics here, which are sufficient for forward flight,

\begin{equation}
\begin{aligned}
    \ddot{X} &= \sum\limits_{j=1}^{N_p} \frac{T_j \sin(\Theta)}{m} - \frac{C_{BD} V_x^2}{m} \\
    \ddot{Z} &= \sum\limits_{j=1}^{N_p} \frac{T_j \cos(\Theta)}{m} - g \\
    \ddot{\Theta} &= \frac{\tau_\Theta}{J_\Theta} = \sum_{j=1}^{N_p} \frac{L_{\Theta,j} T_j}{J_\Theta}
\end{aligned}
\label{eq:2DMotion}
\end{equation}
where $X$ and $Z$ are the horizontal and vertical positions, $V_x$ and $V_z$ are the horizontal and vertical velocities of the vehicle (center of mass) in the global frame,
and $\Theta$ is the pitch angle.
In addition, ${\Sigma}T_{j}$ is the sum of thrusts of all rotors computed by the propeller model (with each rotor indexed by subscript $_j$ and total number of rotors $N_p$), $C_{BD}V_x^2$ 
is the body drag force with $C_{BD}$ being the body drag coefficient, $\tau_\Theta$ is the total pitch-axis torque on the vehicle generated by the thrusts, $J_\Theta$ is the moment of inertia about the pitch axis, and $L_{\Theta,j}$ is the arm length of each rotor thrust to the pitch axis.

Noted that we model the horizontal aerodynamic resistance as a quadratic body-drag term,  
and omit the linear drag component associated with rotor hub and in-plane forces, 
which are known from blade-element theory to grow roughly in proportion to the forward velocity~\cite{Leishmann2000}. 
For the flight regime of interest, i.e., typical mid- to large-scale payload-carrying multirotors operating near energy-optimal velocity,  
the linear drag is comparatively small. 
Aerodynamically, these vehicles operate at low advance
ratio because the rotor-tip speed is much higher than the forward velocity, and hence the rotor-generated in-plane forces scale approximately linearly
with velocity. 
In contrast, the airframe and payload behave as bluff bodies whose drag grows with the square of velocity~\cite{Anderson2017}. 
A drag law that increases faster than linearly is needed for a pronounced non-monotonic EPM-velocity curve typically observed in experiments.
We have performed experimental studies to validate such treatment.
Specifically, we compute the rotor thrust and linear drag for steady forward-flight conditions using a blade-element/momentum model calibrated on thrust-stand data reported in literature \cite{michel2022performance}. 
Then, using real-world flight data, we determine the total drag directly from the measured velocity and pitch angle through steady force balance. 
The body drag term, which is the residual between the total and linear drag, grows approximately with the square of velocity in the neighborhood of the energy optima, and clearly dominates over the linear drag contribution in that regime. 
These physical considerations and experiment results together justify representing the dominant horizontal resistance near the optimal velocity as a quadratic drag term.

Finally, the velocities of each rotor (center) in the vehicle frame can be calculated as
\begin{equation}
\begin{aligned}
    v_{x,j} &= V_x \cos(\Theta) + V_z \sin(\Theta) \\
    v_{z,j} &= -V_x \sin(\Theta) + V_z \cos(\Theta) + \dot{\Theta} x_j
\end{aligned}
\label{eq:Rotorvel}
\end{equation}
where $x_j$ is the x-axis position of the propeller relative to the vehicle center of mass. 

\subsection{Propeller Aerodynamics}
\label{sec:prop_aero}
\begin{figure}[t]
	\centering
		\includegraphics[width=0.35\textwidth]{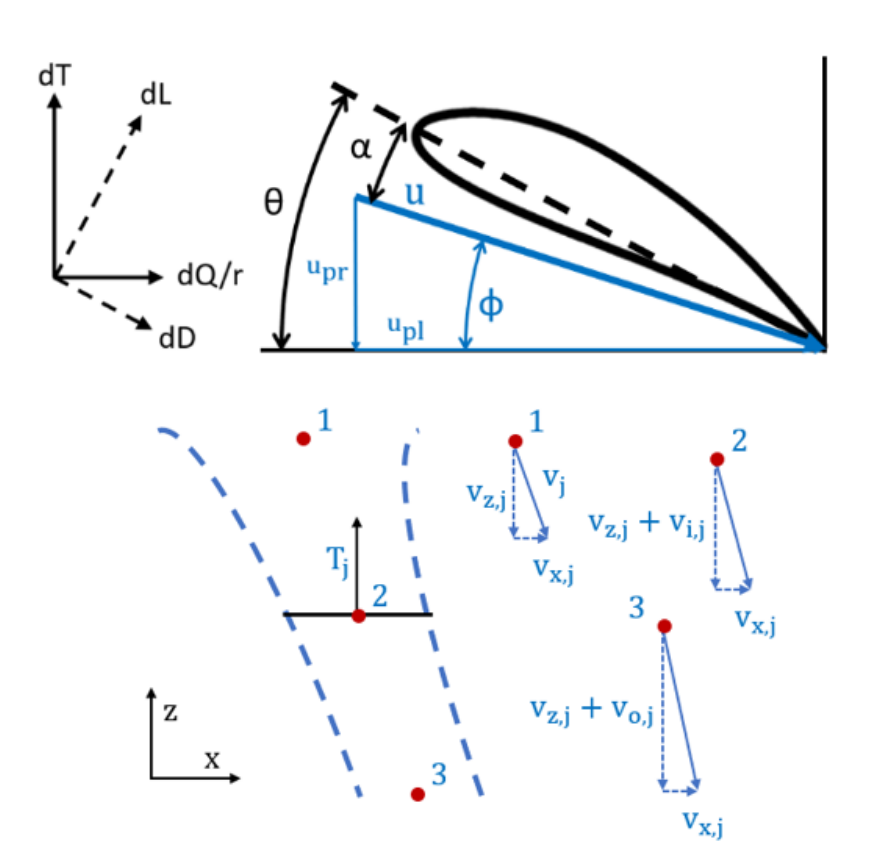}
	\caption{(a) Geometries of a Blade Element Cross-section and (b) Air Flow Momentum of Propeller Disk}
	\label{fig:CrossSection}
\end{figure}

Based on the blade element and momentum theory, the propeller aerodynamic model computes the thrust $T_r$ and torque $Q_r$ generated by a propeller (the subscript $_j$ is omitted in this subsection as all propellers are the same during horizontal flight). 
This is achieved by integrating $dT$ and $dQ$ of propeller blade elements, 
of which the cross section is shown in Fig. \ref{fig:CrossSection}a,
along the blade length coordinate $r$ and averaged over one cycle of the blade rotation angle $\psi$, 
\begin{equation}
\begin{aligned}
	T_r&=\int_{0}^{2\pi}\int_{R_0}^{R}\frac{dT}{2\pi}d{\psi}dr, \, 
     dT = \frac{N_b\rho{u_{pl}}^2ca(\theta-\frac{u_{pr}}{u_{pl}})}{2} \\
	Q_r&= \int_{0}^{2\pi}\int_{R_0}^{R}\frac{dQ}{2\pi} d{\psi}dr, \,
     dQ = \frac{N_b\rho r{u_{pl}}^2c(\phi{a}(\theta-\frac{u_{pr}}{u_{pl}})+c_d)}{2}. 
\end{aligned}
\label{eq:TQsaa}
\end{equation}

In the equation, 
$\rho$ is the air density, $N_b$ is the number of blades per propeller, $c_d$ is the blade drag coefficient, $a$ is the blade lift coefficient factor, $c$ is the blade chord length, $\theta$ is the twist angle of the blade,
and $R$ and $R_0$ are the total and base length of the blade. 
Full details of the derivation with underlying assumptions are available in literature \cite{scaramuzza2022}, \cite{michel2022performance}.
The main idea of Eqn. (\ref{eq:TQsaa}) is to determine the forces based on the dynamic properties of the air inflow to the blade, including the planar inflow velocity along the blade $u_{pl}$, perpendicular inflow velocity $u_{pr}$ vertical to the blade, 
and inflow angle $\phi$, which are shown in Fig. \ref{fig:CrossSection}a. 
These variables can be computed based on the various components of the relative velocity of the air to the blade, including the 
rotation and downwash of the propeller and motion of the whole multicopter, as 
\begin{equation}
\begin{aligned}
	&u_{pl}(r,\psi)=\omega r+v_{x} sin(\psi), \quad 
	u_{pr}(r)=v_i-v_{z} \\
	&\phi(r,\psi)={\tan}^{-1}{(u_{pr}/u_{pl})}. 
\end{aligned}
\label{eq:Rotorinflow}
\end{equation}
Since the
vehicle-frame $z$ axis points upward, so
$v_i-v_z = v_i + V_x\sin\Theta$ is the total inflow through the rotor disc:
the freestream and induced components pass through the disc in the same
direction and therefore add.
$\omega$ is the angular velocity of the propeller rotation.
The downwash effect is captured by the induced air velocity $v_i$ of the rotor disk, 
which is the downward velocity of the air imparted by the propeller.
This velocity needs to be determined by applying the momentum theory to the propeller air stream. 
According to Fig. \ref{fig:CrossSection}b, air enters the propeller airflow at location 1 with a relative velocity \( v \) equal to the total UAV airspeed,
\begin{equation}
\begin{aligned}
	v^2 = {v_x}^2 + {v_z}^2.
\end{aligned}
\label{eq:UAVvel}
\end{equation}
The airflow is accelerated perpendicularly by the propeller disk plane, resulting in an induced velocity \( v_i \) at the disk (location 2) and \( v_o \) at the stream outlet (location 3), both assumed to be uniform across relevant cross-sections. 
By applying the principles of momentum and kinetic energy, \( v_o \) is determined to be twice of \( v_i \). Subsequently, the relationship between the thrust \( T \) and the induced velocity $v_i$ can be established through the momentum principle \cite{textbook},
\begin{equation}
\begin{aligned}
    T_r & = \dot{m}_{air} v_o = 2\dot{m}_{air} v_i
\end{aligned}
\label{eq:MomentumTh}
\end{equation}
where \(\dot{m}_{air} \) is the air mass flow rate at the propeller, 
\begin{equation}
  \dot{m}_{air} = \rho \pi R^2 \sqrt{v_x^2 + (v_i - v_z)^2}.
\label{m_air}
\end{equation}
Furthermore, by substituting ${v_x}$ and ${v_z}$ from the rigid body dynamics relationship in Eqn. \ref{eq:Rotorvel} and applying the steady-state horizontal flight conditions with ${V_z}$ = $\dot{\Theta}$ = 0,
a quartic equation is obtained for solving $v_i$
\begin{equation}
\begin{aligned}
    {v_i}^4 + 2{V_x\sin\Theta}{v_i}^3 + {V_x^2}{v_i}^2 = \Big(\frac{T_r}{{2\rho \pi {R^2}}}\Big)^2,
\end{aligned} 
\label{eq:RearrangedMometumTh}
\end{equation}
where $T_r$ is the thrust of a single rotor, so that the total thrust is $T = N_p T_r$. 
Since all rotors are identical and equally loaded in steady horizontal flight, $T_r$ is the same for every rotor.

\subsection{Energy and Power Consumption}
Power consumption of the multicopter is computed as the sum of the mechanical power of each propeller
\begin{equation}
   P = \sum_j^{N_p}Q_{r,j}\omega_j,
\end{equation}
which is integrated over time to obtain energy consumption.
Since all propellers operate identically in steady horizontal flight, this reduces to $P = N_p Q_r \omega = Q \omega$, 
where $Q_r$ is the torque of a single rotor and $Q = N_p Q_r$ is the total rotor torque. 

\section{Model-Based Analysis of Energy Efficiency}
\label{sec:ModelAnalysis}

We first use the model of Sec.~II to examine how vehicle mass affects
energy-optimal steady horizontal flight in simulation. 
Under $\dot {V}_x = 0$, $V_z = 0$, $\dot{\Theta}=0$,
the metric for energy efficiency is the energy consumption per unit distance,
\begin{equation}
EPM=\frac{P}{V_x}=\frac{Q\omega}{V_x}.
\label{eq:EPMequation}
\end{equation}
We consider mass-only variation of a fixed vehicle configuration, with its
geometry, aerodynamic characteristics, and rotor configuration unchanged.

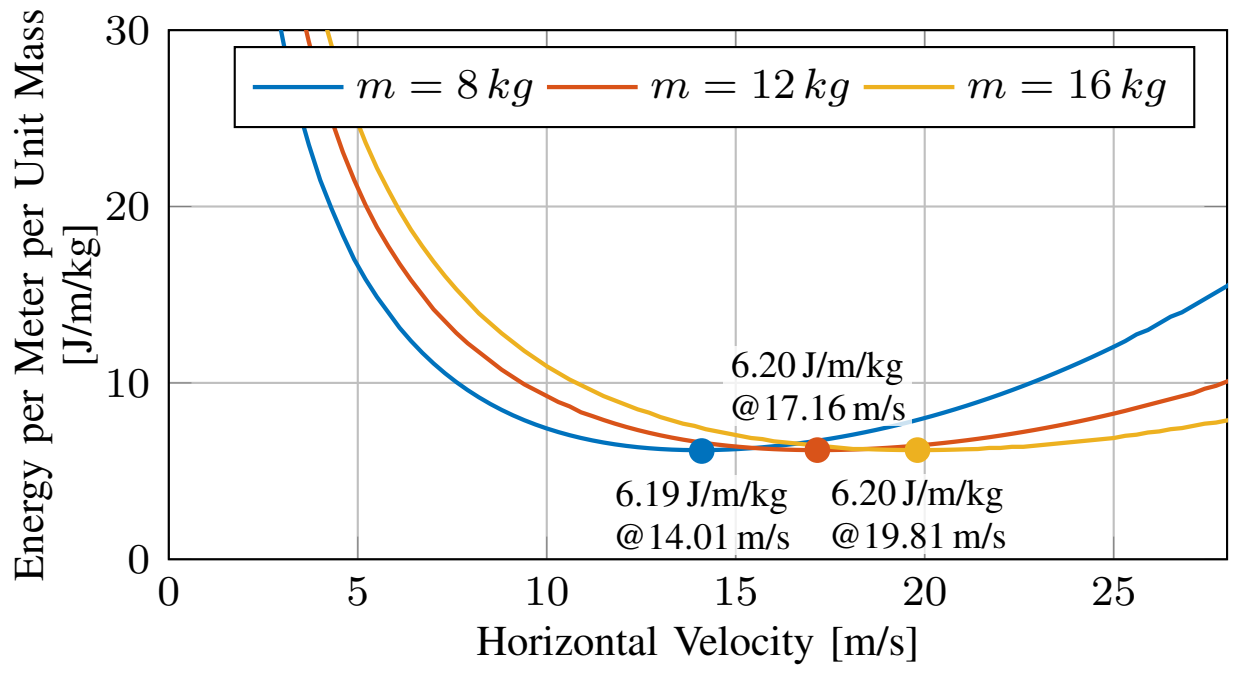
\begin{figure}[ht]
    \centering
    \begin{tikzpicture}[>=stealth]
    \tikzstyle{every node}=[font=\footnotesize]
    \begin{axis}[
        width = 7cm,
        height = 3.5cm,
        xlabel = {Horizontal Velocity [m/s]},
        ylabel = {Energy per Meter per Unit Mass \\ \null [J/m/kg]},
        ylabel style = {align=center},
        xlabel shift=-3pt,
        ylabel shift=-6pt,
        legend columns = 3,
        legend cell align={left},
        legend pos = north east,
        xmin=0.0,
        xmax=28.0,
        ymin=0.0,
        ymax=30,
        xmajorgrids,
        ymajorgrids,
        scale only axis,
        cycle list={{matlab1},{matlab2},{matlab3},{matlab4},{matlab9},{matlab5},{matlab6},{matlab7},{matlab8}},
        ]
    \addplot+[thick, each nth point=100, filter discard warning=false, unbounded coords=discard] table [x=Vx, y expr=\thisrow{epm_8kg}/8, col sep=comma ]{media/epm_vs_mass.csv};
    \addplot+[thick, each nth point=100, filter discard warning=false, unbounded coords=discard] table [x=Vx, y expr=\thisrow{epm_12kg}/12, col sep=comma ]{media/epm_vs_mass.csv};
    \addplot+[thick, each nth point=100, filter discard warning=false, unbounded coords=discard] table [x=Vx, y expr=\thisrow{epm_16kg}/16, col sep=comma ]{media/epm_vs_mass.csv};
    \addlegendentry{$m=8\,kg$}
    \addlegendentry{$m=12\,kg$}
    \addlegendentry{$m=16\,kg$}
    \tikzstyle{opt}=[circle, inner sep=0.06cm, fill]
    \tikzstyle{opt_label}=[fill=white, fill opacity=0.8, text opacity = 1.0, outer sep = 2pt, inner sep = 1pt, align = center, font=\scriptsize]
    \node (epm_opt_8) at (axis cs: 14.1, 49.1/8) [matlab1, opt] {};
    \node (epm_opt_12) at (axis cs: 17.16, 74.26/12) [matlab2, opt] {};
    \node (epm_opt_16) at (axis cs: 19.81, 99.02/16) [matlab3, opt] {};
    \node [opt_label, below = 0mm of epm_opt_8, anchor = north] {6.19\,J/m/kg \\ @14.01\,m/s};
    \node [opt_label, above = 0mm of epm_opt_12, anchor = south] {6.20\,J/m/kg \\ @17.16\,m/s};
    \node [opt_label, below = 0mm of epm_opt_16, anchor = north] {6.20\,J/m/kg \\ @19.81\,m/s};
    \end{axis}  
    
\end{tikzpicture}
    \caption{EPM per unit mass vs Horizontal Velocity under Different Vehicle Mass from Model Analysis}
    \label{fig:EPMbymvsVel}
\end{figure}

Fig.~\ref{fig:EPMbymvsVel} shows the simulated EPM per unit mass versus velocity curves for three
vehicle masses. Each curve exhibits an energy-optimal velocity, which shifts
to higher velocity as mass increases. The corresponding optimal operating
points are summarized in Table~\ref{tab:sim_optima}. 
From these results, we report 3 important findings: the normalized optimal velocity $V_x^*/\sqrt{m}$, optimal pitch angle $\Theta^*$, and normalized
minimum energy $EPM^*/m$ remain nearly constant across mass, suggesting the
three scaling relationships
\begin{equation}
V_x^* \propto m^{1/2},
\label{V*}
\end{equation}
\begin{equation}
EPM^* \propto m,
\label{EPM*}
\end{equation}
\begin{equation}
\Theta^* = \mathrm{const}.
\label{Theta*}
\end{equation}

\begin{table}[ht]
\centering
\caption{Simulated energy-optimal operating points}
\label{tab:sim_optima}
\begin{tabular}{ccccc}
\hline
$m$ (kg) & $V_x^*$ (m/s) & $V_x^*/\sqrt{m}$ &
$\Theta^*$ (deg) & $EPM^*/m$ (J/m/kg) \\
\hline
8  & 14.01 & 4.95 & $15.38$ & 6.19 \\
12 & 17.16 & 4.95 & $15.38$ & 6.20 \\
16 & 19.81 & 4.95 & $15.37$ & 6.20 \\
\hline
\end{tabular}
\end{table}

Note that the increase of $V_x^*$ with $m$ has also been
observed in prior numerical studies~\cite{Hwang11092221,DUKKANCI2021102985},
but an explicit mass-scaling relationship was not established. A narrow range
of energy-optimal $\Theta$ was similarly observed in~\cite{perreault2021delivery},
but its relationship with vehicle mass was not identified. More importantly,
the simulation here indicates that the three trends are not independent:
the mass-invariant optimal pitch angle is associated with both the
$\sqrt{m}$ scaling of optimal velocity and the linear scaling of optimal EPM.
In Sec.~IV, we show analytically that all three relationships follow from a
single mass-similarity property of the steady-state equations.

\section{Derivation of Mass Scaling Laws for Energy-Optimal Flight}
\label{sec:Derivation}

In this section, we prove the three mass-scaling relationships for
energy-optimal flight in Eqs.~(\ref{V*})--(\ref{Theta*}).
The derivation rests on a similarity property of the steady-state equations:
at a fixed pitch angle $\Theta$, the horizontal velocity, induced velocity,
and rotor speed all scale with $\sqrt{m}$, and hence their normalized
quantities depend on $\Theta$ but not on mass.
This similarity leads to $P/m^{3/2}=P'(\Theta)$ and
$EPM/m=F(\Theta)$, where $P'$ and $F$ are independent of mass.
Consequently, the pitch angle minimizing EPM is independent of mass, and the
scaling relationships for the optimal velocity and minimum EPM follow directly.

First, according to the force balance in Fig. \ref{fig: OctoFBD},
we have
\begin{equation}
\begin{array}{l}
 {{{C}}_{{{BD}}}}{{{V}}_x^2} = T\sin \Theta, \quad
  T = \frac{{mg}}{{\cos \Theta }},
\end{array}
\label{ForceBalance}
\end{equation}
which gives
 \begin{equation}
 V_x = \sqrt {{\frac{mg}{C_{BD}}\tan \Theta }} = \sqrt{m}V_x'(\Theta),  
 \label{eq:HorizontalVelEq}
\end{equation}
where
\begin{equation}
  V_x'(\Theta) = \sqrt {{\frac{g}{C_{BD}}\tan \Theta }},
\end{equation}
is the desired mass-independent function of $\Theta$. 

Then, based on the Blade Element Momentum Theory in Eqn. (\ref{eq:TQsaa}) and (\ref{eq:Rotorinflow}), thrust $T$ and torque $Q$ can be re-written as
\begin{equation}
T = \left[ B_{T1} \omega^2 + B_{T2} \frac{V_x^2\cos^2\Theta}{2} - B_{T3} \omega (V_x \sin \Theta + v_i) \right]
\label{eq:ThrustEq1}
\end{equation}
with constants
\begin{equation}
\begin{array}{l}
B_{T1} = \frac{N_p N_b \rho a}{2} \int_{R_0}^R r^2 c(r) \theta(r) \, dr \\
B_{T2} = \frac{N_p N_b \rho a}{2} \int_{R_0}^R c(r) \theta(r) \, dr, \quad
B_{T3} = \frac{N_p N_b \rho a}{2} \int_{R_0}^R r c(r) \, dr
\end{array}
\label{eq:ThrustEqCoeff}
\end{equation}
and
\begin{equation}
\begin{split}
Q &= C_{Q1} \omega^2 + C_{Q2} V_x^2\cos^2\Theta + C_{Q3} (V_x \sin \Theta + v_i) \omega \\
  &\quad - C_{Q4} (V_x \sin \Theta + v_i)^2
\end{split}
\label{eq:TorqueEq1}
\end{equation}
with constants 
\begin{equation}
\begin{array}{l}
C_{Q1} = \frac{N_p N_b \rho c_d}{2} \int_{R_0}^R r^3 c(r) \, dr, \quad
C_{Q2} = \frac{N_p N_b \rho c_d}{4} \int_{R_0}^R r c(r) \, dr, \\[4pt]
C_{Q3} = B_{T1}, \quad C_{Q4} = B_{T3}.
\end{array}
\label{eq:TorqueEqCoeff}
\end{equation}
Note that the thrust and torque of all propellers are equal under the steady-state flight, giving 
$T = N_p T_r$.

We now state the similarity property on which the three scaling laws rest.

\begin{lemma}[Mass similarity of steady-state rotor solution]
\label{lem:similarity}
Consider steady level flight of a multicopter at pitch angle
$\Theta \in (0,\pi/2)$, with the body drag coefficient $C_{BD}$, the rotor geometry and blade coefficients, and the air density held constant as the vehicle mass $m$
varies. Define the non-dimensional induced velocity, tip-speed ratio and rotor thrust coefficient
\begin{equation}
\bar{v}_i = \frac{v_i}{V_x}, \qquad
\bar{\omega} = \frac{\omega R}{V_x}, \qquad
C_T = \frac{T_r}{\rho \pi R^2 V_x^2}.
\label{eq:nondim}
\end{equation}
Then $C_T$, $\bar{v}_i$ and $\bar{\omega}$ are functions of $\Theta$ only and are independent of $m$. 
Consequently, we have 
\begin{equation}
v_i = \sqrt{m}\, v_i'(\Theta) \quad v_i'(\Theta) = V_x'(\Theta)\bar{v}_i(\Theta)
\label{eq: v_i}
\end{equation}
\begin{equation}
\omega = \sqrt{m}\,\omega'(\Theta), \quad
\omega'(\Theta) = V_x'(\Theta)\bar{\omega}(\Theta)/R.
\label{eq:omega} 
\end{equation}
\end{lemma}

\begin{proof}
The force balance (\ref{ForceBalance}) gives $T_r = mg/(N_p\cos\Theta)$ and
$V_x^2 = (mg/C_{BD})\tan\Theta$. Hence their ratio
\begin{equation}
C_T = \frac{mg/(N_p \cos\Theta)}{\rho \pi R^2 (mg/C_{BD})\tan\Theta}
    = \frac{C_{BD}}{N_p \rho \pi R^2 \sin\Theta},
\label{eq:CT}
\end{equation}
in which the mass cancels identically. This is the single fact from which everything else follows.

Dividing the momentum-theory quartic (\ref{eq:RearrangedMometumTh}) by $V_x^4$ and
using (\ref{eq:nondim}) yield
\begin{equation}
\bar{v}_i^{\,2}\big(\bar{v}_i^{\,2} + 2\sin\Theta\,\bar{v}_i + 1\big)
= \Big(\frac{C_T}{2}\Big)^{\!2},
\label{eq:quartic_nd}
\end{equation}
whose coefficients only involve $\Theta$ (and constants) by (\ref{eq:CT}). Hence the solution 
$\bar{v}_i(\Theta)$ only depends on $\Theta$. 

Dividing the thrust relation (\ref{eq:ThrustEq1}) by $V_x^2$, substituting
$T = mg/\cos\Theta$ and $V_x^2 = (mg/C_{BD})\tan\Theta$, and using (\ref{eq:nondim}) give 
\begin{equation}
\frac{B_{T1}}{R^2}\bar{\omega}^2 - \frac{B_{T3}}{R}\bar{\omega}\big(\sin\Theta + \bar{v}_i\big) + \frac{B_{T2}}{2}\cos^2\Theta = \frac{C_{BD}}{\sin\Theta},
\label{eq:quad_nd}
\end{equation}
whose coefficients again involve only $\Theta$. Hence the solution is written as $\bar{\omega}(\Theta)$.
Since $V_x = \sqrt{m}\,V_x'(\Theta)$ by (\ref{eq:HorizontalVelEq}), (\ref{eq: v_i}) and (\ref{eq:omega}) follow directly from the definitions in (\ref{eq:nondim}).
\end{proof}

Next, the torque $Q$ from Eqn. (\ref{eq:TorqueEq1}) can be re-written by substituting the expressions for $V_x$ in Eq. (\ref{eq:HorizontalVelEq}), $v_i$ in Eq. (\ref{eq: v_i}), and $\omega$ in Eq. (\ref{eq:omega}) as
\begin{equation}
\begin{aligned}
Q &= C_{Q1}m\,\omega'^2(\Theta) + C_{Q2}m\,V_x'^2(\Theta)\cos^2\Theta \\
  &\quad + C_{Q3}m\big(V_x'(\Theta)\sin\Theta + v_i'(\Theta)\big)\omega'(\Theta) \\
  &\quad - C_{Q4}m\big(V_x'(\Theta)\sin\Theta + v_i'(\Theta)\big)^2 .
\end{aligned}
\label{eq:Q_sub}
\end{equation}
Note that since $V_x'$, $v_i'$, and $\omega'$ are all functions of only $\Theta$, we can write $Q$ as
\begin{equation}
 Q = m Q'(\Theta).
\label{eq:Q} 
\end{equation}
Eqn. (\ref{eq:Q}) follows from the fact that every term of (\ref{eq:TorqueEq1}) is quadratic in the velocities, each of which scales as $\sqrt{m}$ by Lemma~\ref{lem:similarity}. It therefore holds irrespective of the numerical values of the blade constants.

Since the rotor torque scales as $Q \propto m$ by Eqn. (\ref{eq:Q}) and the rotor speed as $\omega \propto \sqrt{m}$ by (\ref{eq:omega}), the power required in steady horizontal flight satisfies
\begin{equation}
P(\Theta;m) = Q\omega = m^{3/2}P'(\Theta), \quad
P'(\Theta) = Q'(\Theta)\,\omega'(\Theta).
\label{eq:P_scaling}
\end{equation}
Combining (\ref{eq:P_scaling}) with $V_x(\Theta;m) = \sqrt{m}\,V_x'(\Theta)$ from (\ref{eq:HorizontalVelEq}), the mass-normalized energy per meter in Eqn. (\ref{eq:EPMequation}) becomes
\begin{equation}
\frac{EPM(\Theta;m)}{m} = \frac{P}{m V_x}
= \frac{m^{3/2}P'(\Theta)}{m\sqrt{m}\,V_x'(\Theta)}
= \frac{P'(\Theta)}{V_x'(\Theta)} \triangleq F(\Theta),
\label{eq:EPMequation2}
\end{equation}
where $F$ is independent of mass for a fixed vehicle configuration. 
Therefore, the optimal pitch angle 
\begin{equation}
\Theta^*=\arg\min_{\Theta}EPM(\Theta;m)
=\arg\min_{\Theta}F(\Theta),
\label{eq:ThetaStarArgmin}
\end{equation}
is independent of mass, proving Eqn.~(\ref{Theta*}).
Evaluating Eqns.~(\ref{eq:HorizontalVelEq}) and
(\ref{eq:EPMequation2}) at this common $\Theta^*$ gives
\begin{equation}
V_x^*=\sqrt{m}\,V_x'(\Theta^*)\propto m^{1/2},
\label{eq:VstarScaling}
\end{equation}
and
\begin{equation}
EPM^*=mC, \qquad
C=F(\Theta^*)=\frac{Q'(\Theta^*)\omega'(\Theta^*)}{V_x'(\Theta^*)},
\label{eq: C}
\end{equation}
proving Eqns.~(\ref{V*}) and~(\ref{EPM*}), respectively.

\begin{remark}[Why the laws hold]
\label{rem:similarity}
Suppose the vehicle mass is multiplied by a factor $k$. Every force in Fig.~\ref{fig: OctoFBD} must then also be multiplied by $k$. Every aerodynamic force, 
i.e. body drag, rotor thrust and torque, is quadratic in velocity, so multiplying every velocity by $\sqrt{k}$ multiplies every force by $k$ while leaving every angle unchanged: the pitch angle $\Theta$, the inflow angle $\phi$ of (\ref{eq:Rotorinflow}), and the blade angle of attack are all ratios of velocities and are therefore invariant. Heavier vehicles fly through a geometrically similar flow field simply faster. Mass sets the scale of the steady-state solution but not its shape. The three scaling laws are the three consequences of this similarity: angles are invariant, so the energy-optimal pitch angle is invariant; velocities scale as $\sqrt{m}$, so the optimal cruise speed scales as $\sqrt{m}$; and power is force times velocity and therefore scales as $m^{3/2}$, while energy per meter is power divided by velocity and therefore scales linearly in mass.
\end{remark}

\section{Experimental Validation}
\label{sec:Experimental}

Outdoor flight tests were conducted on two multicopter platforms with different
sizes, rotor counts, and propulsion systems at the Woodland-Davis Aeromodelers field. 
The octorotor is based on a DJI Spreading Wings S1000 equipped with a Pixhawk 2.1
flight controller and onboard GNSS, current, and voltage measurements, as shown
in Fig.~\ref{fig:Overall}a. The quadrotor is a Hexsoon EDU-650 V2 equipped with
 a CubeOrange flight controller and similar onboard instrumentation. 

The octorotor was tested at total masses of 7, 8, and 9~kg, spanning a 29\%
mass variation, while the quadrotor was tested at 5 masses
from 3.58 to 5.17~kg, spanning 44\%. Added payload was mounted compactly 
so that the projected area and overall aerodynamic shape
remained approximately unchanged, consistent with the mass-only variation
assumed in Sec.~\ref{sec:Derivation}. Velocity set points were flown in both
directions along a reciprocal straight-line course to mitigate directional wind
effects through averaging. The campaigns are summarized in
Table~\ref{tab:campaign}.
Power, horizontal velocity, and pitch angle were recorded onboard. Steady-flight
segments were identified by requiring the ground speed to remain near its
commanded set point for at least 3~s; all reported
quantities were averaged over these segments. EPM was computed as
$\int P\,dt/\int V_x\,dt$. Reciprocal measurements were averaged within each
mass before cross-mass comparison. The pitch angle was obtained from the onboard
flight-controller attitude estimate rather than inferred from the force-balance
model.

\begin{table}[ht]
\centering
\caption{Summary of experimental data}
\label{tab:campaign}
\begin{tabular}{lccccc}
\hline
Platform & Masses & Range (kg) & Variation & Set points & Steady legs \\
\hline
Quadrotor & 5 & 3.58--5.17 & 44\% & 6 & 55 \\
Octorotor & 3 & 7.00--9.00 & 29\% & 6--7 & 39 \\
\hline
\end{tabular}
\end{table}

\subsection{Full-Range Validation of Similarity Relations}

We first validate the scaling laws through the full-range similarity relationships.
According to Eqs.~(\ref{eq:HorizontalVelEq}) and~(\ref{eq:P_scaling}), we have $V_x/\sqrt{m}=V_x'(\Theta)$ and
$P/m^{3/2}=P'(\Theta)$, which directly lead to
$EPM/m=F(\Theta)$ and hence the three optimal-flight scaling laws.
We therefore examine whether $V_x/\sqrt{m}$ and $P/m^{3/2}$ over $\Theta$ are invariant across different masses. 
We also validate the scaling laws
using the experimentally identified optimal operating points in the next
subsection. However, validation based on the optima requires the EPM minimum to
be experimentally resolved at each mass, whereas the full-range approach uses
the complete operating range including mass cases for which the optimum cannot be resolved.

Figs.~\ref{fig:Pnorm} and~\ref{fig:Vnorm} show the
$P/m^{3/2}$ and $V_x/\sqrt{m}$ versus $\Theta$ for all mass cases of each platform.
The dashed curves show the corresponding model-based fits obtained from the steady-state dynamic and BEMT relations in Sec. II, 
with platform-specific parameters fitted to the experimental data.
\begin{figure}[ht]
    \centering
    \subfloat[Quadrotor]{\scalebox{0.95}{\begin{tikzpicture}[>=stealth]
    \tikzstyle{every node}=[font=\scriptsize]
    \begin{axis}[
        width = 7cm,
        height = 3.0cm,
        xlabel = {Pitch angle, $\Theta$ [deg]},
        ylabel = {$P/m^{3/2}$\\ \null [W kg$^{-3/2}$]},
        ylabel style = {align=center},
        xlabel shift=-4pt,
        tick label style={font=\scriptsize},
        major grid style={black!12},
        ylabel shift=-6pt,
        legend columns = 3,
        legend cell align={left},
        legend style={font=\scriptsize, row sep=-1pt, draw=black!45, fill=white,
            at={(0.5,1.04)}, anchor=south, inner sep=2pt,
            /tikz/every even column/.append style={column sep=4pt}},
        xmin=6, xmax=23,
        ymin=45, ymax=88,
        ytick={50,60,70,80},
        xmajorgrids,
        ymajorgrids,
        scale only axis,
        ]
    \addplot[only marks, mark=*, mark size=1.6pt, color=matlab1, unbounded coords=discard, filter discard warning=false] table [x=th_m3p584, y=Pn_m3p584, col sep=comma]{media/norm_quad.csv};
    \addlegendentry{3.58 kg}
    \addplot[only marks, mark=square*, mark size=1.6pt, color=matlab2, unbounded coords=discard, filter discard warning=false] table [x=th_m3p826, y=Pn_m3p826, col sep=comma]{media/norm_quad.csv};
    \addlegendentry{3.83 kg}
    \addplot[only marks, mark=triangle*, mark size=1.6pt, color=matlab5, unbounded coords=discard, filter discard warning=false] table [x=th_m4p242, y=Pn_m4p242, col sep=comma]{media/norm_quad.csv};
    \addlegendentry{4.24 kg}
    \addplot[only marks, mark=diamond*, mark size=1.6pt, color=matlab3, unbounded coords=discard, filter discard warning=false] table [x=th_m4p666, y=Pn_m4p666, col sep=comma]{media/norm_quad.csv};
    \addlegendentry{4.67 kg}
    \addplot[only marks, mark=pentagon*, mark size=1.6pt, color=matlab4, unbounded coords=discard, filter discard warning=false] table [x=th_m5p168, y=Pn_m5p168, col sep=comma]{media/norm_quad.csv};
    \addlegendentry{5.17 kg}
    \addplot[black, thick, dashed, mark=none] table [x=th, y=Pn, col sep=comma]{media/bemt_quad.csv};
    \addlegendentry{BEMT model fit}
    \end{axis}
\end{tikzpicture}}} \\
    \subfloat[Octorotor]{\scalebox{0.95}{\begin{tikzpicture}[>=stealth]
    \tikzstyle{every node}=[font=\scriptsize]
    \begin{axis}[
        width = 7cm,
        height = 3.0cm,
        xlabel = {Pitch angle, $\Theta$ [deg]},
        ylabel = {$P/m^{3/2}$\\ \null [W kg$^{-3/2}$]},
        ylabel style = {align=center},
        xlabel shift=-4pt,
        tick label style={font=\scriptsize},
        major grid style={black!12},
        ylabel shift=-6pt,
        legend columns = 4,
        legend cell align={left},
        legend style={font=\scriptsize, row sep=-1pt, draw=black!45, fill=white,
            at={(0.5,1.04)}, anchor=south, inner sep=2pt,
            /tikz/every even column/.append style={column sep=4pt}},
        xmin=8, xmax=29,
        ymin=30, ymax=84,
        ytick={40,60,80},
        xmajorgrids,
        ymajorgrids,
        scale only axis,
        ]
    \addplot[only marks, mark=*, mark size=1.6pt, color=matlab1, unbounded coords=discard, filter discard warning=false] table [x=th_m7p0, y=Pn_m7p0, col sep=comma]{media/norm_octo.csv};
    \addlegendentry{7 kg}
    \addplot[only marks, mark=square*, mark size=1.6pt, color=matlab2, unbounded coords=discard, filter discard warning=false] table [x=th_m8p0, y=Pn_m8p0, col sep=comma]{media/norm_octo.csv};
    \addlegendentry{8 kg}
    \addplot[only marks, mark=triangle*, mark size=1.6pt, color=matlab5, unbounded coords=discard, filter discard warning=false] table [x=th_m9p0, y=Pn_m9p0, col sep=comma]{media/norm_octo.csv};
    \addlegendentry{9 kg}
    \addplot[black, thick, dashed, mark=none] table [x=th, y=Pn, col sep=comma]{media/bemt_octo.csv};
    \addlegendentry{BEMT model fit}
    \end{axis}
\end{tikzpicture}}}
    \caption{$P/m^{3/2}$ versus $\Theta$ for all mass
    configurations of each platform.}
    \label{fig:Pnorm}
\end{figure}
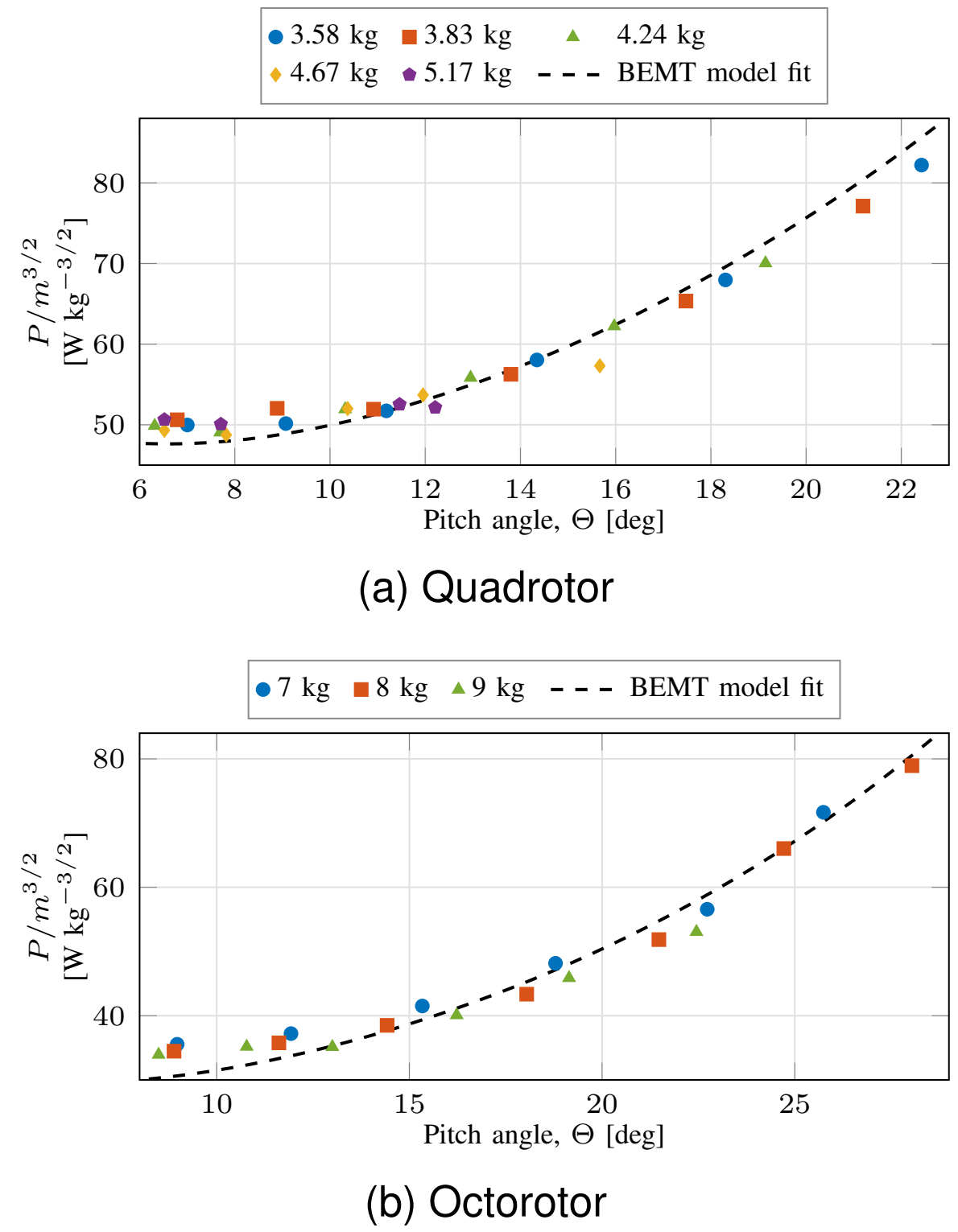
\begin{figure}[ht]
    \centering
    \subfloat[Quadrotor]{\scalebox{0.95}{\begin{tikzpicture}[>=stealth]
    \tikzstyle{every node}=[font=\scriptsize]
    \begin{axis}[
        width = 7cm,
        height = 3.0cm,
        xlabel = {Pitch angle, $\Theta$ [deg]},
        ylabel = {$V_x/\sqrt{m}$\\ \null [m s$^{-1}$kg$^{-1/2}$]},
        ylabel style = {align=center},
        xlabel shift=-4pt,
        tick label style={font=\scriptsize},
        major grid style={black!12},
        ylabel shift=-6pt,
        legend columns = 3,
        legend cell align={left},
        legend style={font=\scriptsize, row sep=-1pt, draw=black!45, fill=white,
            at={(0.5,1.04)}, anchor=south, inner sep=2pt,
            /tikz/every even column/.append style={column sep=4pt}},
        xmin=6, xmax=23,
        ymin=3.0, ymax=10.0,
        ytick={4,6,8,10},
        xmajorgrids,
        ymajorgrids,
        scale only axis,
        ]
    \addplot[only marks, mark=*, mark size=1.6pt, color=matlab1, unbounded coords=discard, filter discard warning=false] table [x=th_m3p584, y=Vn_m3p584, col sep=comma]{media/norm_quad.csv};
    \addlegendentry{3.58 kg}
    \addplot[only marks, mark=square*, mark size=1.6pt, color=matlab2, unbounded coords=discard, filter discard warning=false] table [x=th_m3p826, y=Vn_m3p826, col sep=comma]{media/norm_quad.csv};
    \addlegendentry{3.83 kg}
    \addplot[only marks, mark=triangle*, mark size=1.6pt, color=matlab5, unbounded coords=discard, filter discard warning=false] table [x=th_m4p242, y=Vn_m4p242, col sep=comma]{media/norm_quad.csv};
    \addlegendentry{4.24 kg}
    \addplot[only marks, mark=diamond*, mark size=1.6pt, color=matlab3, unbounded coords=discard, filter discard warning=false] table [x=th_m4p666, y=Vn_m4p666, col sep=comma]{media/norm_quad.csv};
    \addlegendentry{4.67 kg}
    \addplot[only marks, mark=pentagon*, mark size=1.6pt, color=matlab4, unbounded coords=discard, filter discard warning=false] table [x=th_m5p168, y=Vn_m5p168, col sep=comma]{media/norm_quad.csv};
    \addlegendentry{5.17 kg}
    \addplot[black, thick, dashed, mark=none] table [x=th, y=Vn, col sep=comma]{media/cbd_quad.csv};
    \addlegendentry{Eq.~(\ref{eq:HorizontalVelEq}), $C_{BD}=0.0518$}
    \end{axis}
\end{tikzpicture}}} \\
    \subfloat[Octorotor]{\scalebox{0.95}{\begin{tikzpicture}[>=stealth]
    \tikzstyle{every node}=[font=\scriptsize]
    \begin{axis}[
        width = 7cm,
        height = 3.0cm,
        xlabel = {Pitch angle, $\Theta$ [deg]},
        ylabel = {$V_x/\sqrt{m}$\\ \null [m s$^{-1}$kg$^{-1/2}$]},
        ylabel style = {align=center},
        xlabel shift=-4pt,
        tick label style={font=\scriptsize},
        major grid style={black!12},
        ylabel shift=-6pt,
        legend columns = 4,
        legend cell align={left},
        legend style={font=\scriptsize, row sep=-1pt, draw=black!45, fill=white,
            at={(0.5,1.04)}, anchor=south, inner sep=2pt,
            /tikz/every even column/.append style={column sep=4pt}},
        xmin=8, xmax=29,
        ymin=2.9, ymax=7.6,
        ytick={3,5,7},
        xmajorgrids,
        ymajorgrids,
        scale only axis,
        ]
    \addplot[only marks, mark=*, mark size=1.6pt, color=matlab1, unbounded coords=discard, filter discard warning=false] table [x=th_m7p0, y=Vn_m7p0, col sep=comma]{media/norm_octo.csv};
    \addlegendentry{7 kg}
    \addplot[only marks, mark=square*, mark size=1.6pt, color=matlab2, unbounded coords=discard, filter discard warning=false] table [x=th_m8p0, y=Vn_m8p0, col sep=comma]{media/norm_octo.csv};
    \addlegendentry{8 kg}
    \addplot[only marks, mark=triangle*, mark size=1.6pt, color=matlab5, unbounded coords=discard, filter discard warning=false] table [x=th_m9p0, y=Vn_m9p0, col sep=comma]{media/norm_octo.csv};
    \addlegendentry{9 kg}
    \addplot[black, thick, dashed, mark=none] table [x=th, y=Vn, col sep=comma]{media/cbd_octo.csv};
    \addlegendentry{Eq.~(\ref{eq:HorizontalVelEq}), $C_{BD}=0.1120$}
    \end{axis}
\end{tikzpicture}}}
    \caption{$V_x/\sqrt{m}$ versus $\Theta$ for all mass
    configurations of each platform.}
    \label{fig:Vnorm}
\end{figure}
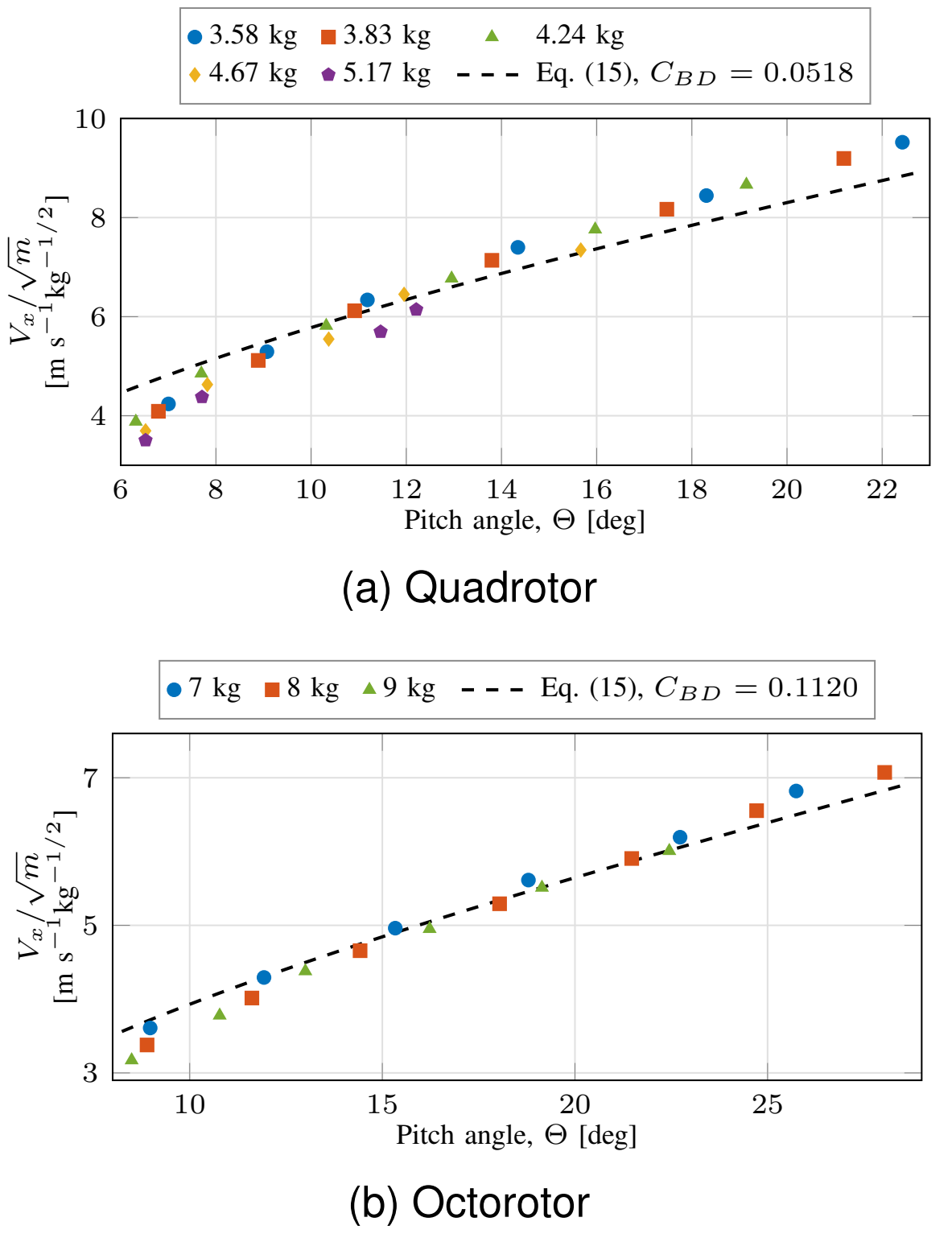
It is seen that both relationships show a strong collapse across mass. 
To quantify the agreement, the measurement data for each mass are interpolated
to common values of $\Theta$, and the cross-mass
coefficients of variation (CoV) are computed.
For quadrotor, the cross-mass
CoV is 0.80--1.88\% for $P/m^{3/2}$ and 1.52--4.68\% for
$V_x/\sqrt{m}$; 
for octorotor, the CoVs are 2.07--4.04\% and 1.53--3.89\%, respectively.
The small CoVs indicate that $P/m^{3/2}$ and
$V_x/\sqrt{m}$ remain approximately mass-independent functions of $\Theta$, 
directly validating the similarity relationships underlying the three optimal-flight scaling laws.

\subsection{Validation of Energy Optima}

\begin{figure}[ht]
    \centering
\begin{tikzpicture}

\begin{groupplot}[
  group style={
    group size=1 by 2,
    vertical sep=2.2cm,
  },
  width=0.88\linewidth,
  height=0.43\linewidth,
  axis lines=box,
  grid=major,
  major grid style={black!12},
  tick align=outside,
  tick label style={font=\footnotesize},
  label style={font=\small},
  ylabel style={align=center, yshift=-2pt},
  xlabel style={yshift=2pt},
  xlabel={Horizontal velocity, $V_x$ [m\,s$^{-1}$]},
  ylabel={EPM per mass\\ $[$J\,m$^{-1}$kg$^{-1}]$},
  legend style={draw=black!45, fill=white, font=\scriptsize,
    /tikz/every even column/.append style={column sep=5pt},},
]

\nextgroupplot[
  xmin=7.2, xmax=18.8,
  ymin=7.4, ymax=15.0,
  xtick={8,10,12,14,16,18},
  ytick={8,10,12,14},
  legend to name=quadlegend,
  legend columns=5,
]

\addplot[only marks, mark=o, mark size=2.0pt, matlab1, mark options={fill=white}] coordinates {
  (8.0241,11.7905) (10.0202,9.4758) (11.9987,8.1611) (14.0072,7.8464) (15.9858,8.0486) (18.0231,8.6363)
};
\addlegendentry{3.58 kg}
\addplot[matlab1, thick, samples=200, domain=8.0241:18.0231, forget plot] {(-0.0053763142)*x^3+(0.3020974)*x^2+(-5.3148607)*x+(37.773578)};

\addplot[only marks, mark=square, mark size=2.0pt, matlab2, mark options={fill=white}] coordinates {
  (8.0004,12.3762) (10.0083,10.1723) (11.9687,8.4868) (13.9606,7.8807) (15.9772,8.0035) (17.9821,8.3873)
};
\addlegendentry{3.83 kg}
\addplot[matlab2, thick, samples=200, domain=8.0004:17.9821, forget plot] {(-0.0026958189)*x^3+(0.1956811)*x^2+(-4.0525566)*x+(33.696528)};

\addplot[only marks, mark=triangle, mark size=2.2pt, matlab5, mark options={fill=white}] coordinates {
  (7.9957,12.8500) (9.9992,10.1075) (11.9808,8.9286) (13.9468,8.2448) (15.9857,8.0227) (17.8466,8.0951)
};
\addlegendentry{4.24 kg}
\addplot[matlab5, thick, samples=200, domain=7.9957:17.8466, forget plot] {(-0.0075133093)*x^3+(0.37287529)*x^2+(-6.1725582)*x+(42.171798)};

\addplot[only marks, mark=diamond, mark size=2.2pt, matlab4, mark options={fill=white}] coordinates {
  (7.9835,13.3436) (10.0004,10.5273) (11.9820,9.3783) (13.9353,8.3330) (15.8671,7.8087)
};
\addlegendentry{4.67 kg}
\addplot[matlab4, thick, samples=200, domain=7.9835:15.8671, forget plot] {(-0.01190029)*x^3+(0.50982732)*x^2+(-7.6000486)*x+(47.55328)};

\addplot[only marks, mark=pentagon, mark size=2.2pt, matlab7, mark options={fill=white}] coordinates {
  (7.9659,14.4516) (9.9509,11.4390) (12.9434,9.2267) (13.9640,8.4910)
};
\addlegendentry{5.17 kg}
\addplot[matlab7, thick, samples=200, domain=7.9659:13.9640, forget plot] {(0.11482347)*x^2+(-3.4815369)*x+(34.845723)};

\addplot[only marks, mark=*, mark size=2.3pt, black, forget plot] coordinates {
  (14.1150,7.8230) (15.0118,7.8380)
};

\nextgroupplot[
  xmin=8.8, xmax=20.7,
  ymin=7.8, ymax=11.5,
  xtick={10,12,14,16,18,20},
  ytick={8,9,10,11},
  legend to name=octolegend,
  legend columns=3,
]

\addplot[only marks, mark=o, mark size=2.0pt, matlab1, mark options={fill=white}] coordinates {
  (9.5487,9.8477) (11.3595,8.6634) (13.1274,8.3648) (14.8516,8.5780) (16.3878,9.1324) (18.0448,10.5182)
};
\addlegendentry{7 kg}
\addplot[matlab1, thick, samples=200, domain=9.5487:18.0448, forget plot] {(-0.00017538465)*x^3+(0.10886037)*x^2+(-2.8221979)*x+(27.00115)};

\addplot[only marks, mark=square, mark size=2.0pt, matlab2, mark options={fill=white}] coordinates {
  (9.5566,10.1986) (11.3558,8.9065) (13.1703,8.2653) (14.9687,8.1900) (16.7041,8.7833) (18.5372,10.0761) (20.0034,11.1629)
};
\addlegendentry{8 kg}
\addplot[matlab2, thick, samples=200, domain=9.5566:20.0034, forget plot] {(-0.0018614891)*x^3+(0.17389408)*x^2+(-3.7745419)*x+(32.034715)};

\addplot[only marks, mark=triangle, mark size=2.2pt, matlab5, mark options={fill=white}] coordinates {
  (9.5160,10.6938) (11.3353,9.3058) (13.1361,8.0182) (14.8551,8.0930) (16.5331,8.3226) (18.0253,8.8353)
};
\addlegendentry{9 kg}
\addplot[matlab5, thick, samples=200, domain=9.5160:18.0253, forget plot] {(-0.002382317)*x^3+(0.19273278)*x^2+(-4.1303871)*x+(34.64815)};

\addplot[only marks, mark=*, mark size=2.3pt, black, forget plot] coordinates {
  (13.3963,8.3087) (14.0002,8.1664) (14.7484,8.0114)
};

\end{groupplot}

\node[anchor=south] at ($(group c1r1.north)+(0,1.5mm)$)
  {\pgfplotslegendfromname{quadlegend}};

\node[anchor=south] at ($(group c1r2.north)+(0,1.5mm)$)
  {\pgfplotslegendfromname{octolegend}};

\end{tikzpicture}
    \caption{Mass-normalized EPM versus horizontal velocity for the quadrotor
    (top) and octorotor (bottom). Curves show least-squares cubic fits to the measured
    set points.}
    \label{fig:epm_vs_vx}
\end{figure}
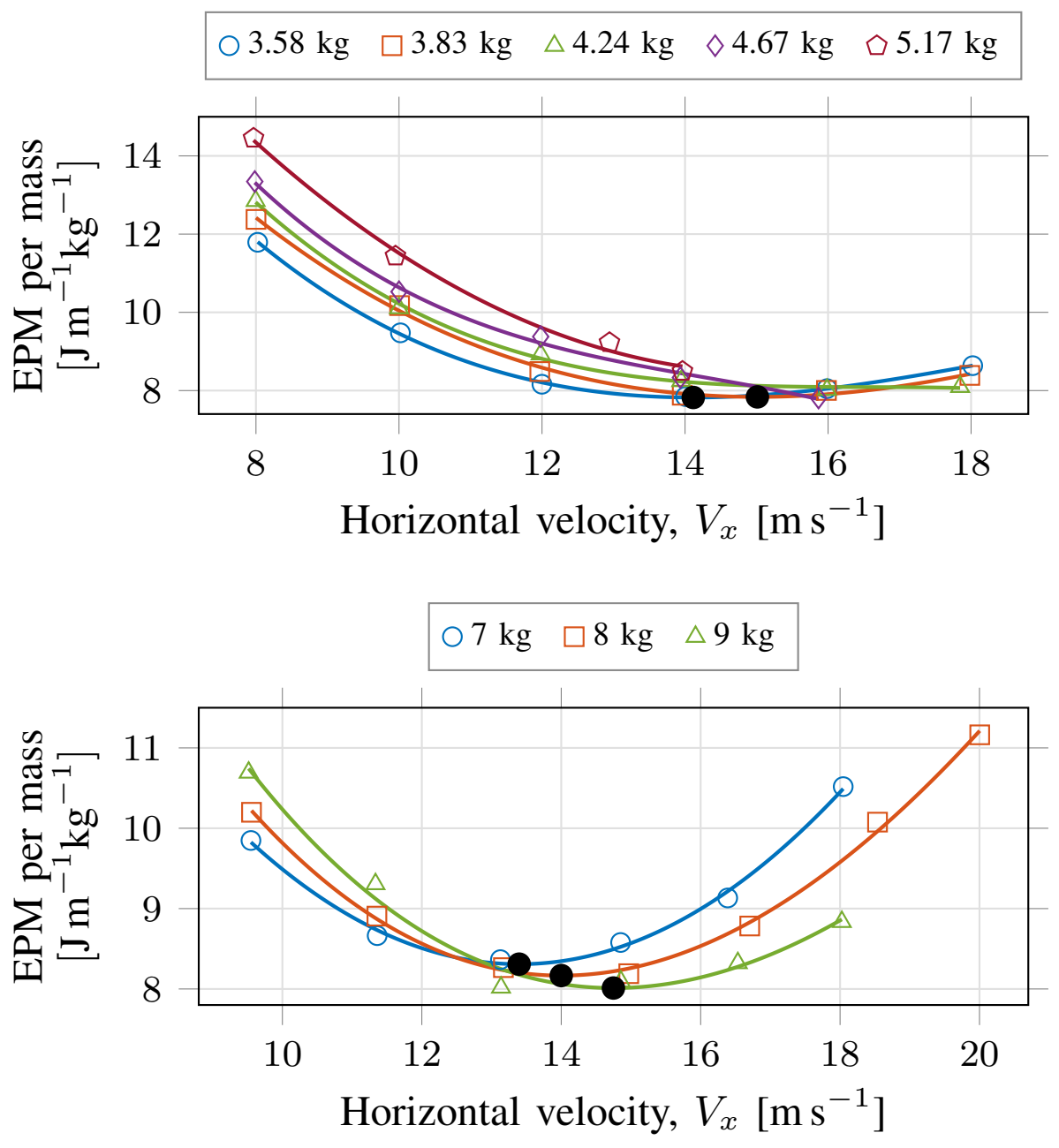

We next validate the scaling laws using the experimentally identified energy-optimal operating points. 
Fig.~\ref{fig:epm_vs_vx} shows the mass-normalized EPM curves. For the heavier mass cases, the velocities required to bracket the EPM minimum exceeded the propulsion capability of the vehicle; consequently, the optimum can be identified for only 5 mass cases, as reported in Table~\ref{tab:optima}. Nevertheless, the full-range analysis in Sec. V-A provides validation across all mass configurations without requiring the EPM minimum to be resolved.

\begin{table}[ht]
\centering
\caption{Experimentally identified energy-optimal operating points}
\label{tab:optima}
\begin{tabular}{lcccc}
\hline
Platform & Mass (kg) & $\Theta^*$ (deg) & $V_x^*/\sqrt{m}$ &
$EPM^*/m$ (J/m/kg) \\
\hline
Octorotor & 7.00 & 16.34 & 5.06 & 8.31 \\
Octorotor & 8.00 & 16.19 & 4.95 & 8.17 \\
Octorotor & 9.00 & 15.50 & 4.92 & 8.01 \\
Quadrotor & 3.58 & 14.30 & 7.46 & 7.82 \\
Quadrotor & 3.83 & 14.82 & 7.67 & 7.84 \\
\hline
\end{tabular}
\end{table}

For the octorotor, the identified $\Theta^*$ varies by only $0.84^\circ$
across the 29\% mass increase, while $V_x^*/\sqrt{m}$ and $EPM^*/m$ have
coefficients of variation of 1.6\% and 1.8\%, respectively. 
Although $EPM^*/m$ shows a slight monotonic decrease with increasing mass from the graph, 
the variation remains small (1.8\%) and represents only a minor deviation from the
predicted mass invariance.
For the two resolved quadrotor cases, $\Theta^*$ differs by $0.52^\circ$,
while $V_x^*/\sqrt{m}$ and $EPM^*/m$ differ by 2.9\% and 0.2\%,
respectively.
These
results are obtained from cubic fits to the measured EPM curves, 
providing consistent experimental validation
for all three scaling laws on both platforms.
To assess the sensitivity to curve fitting used to identify the optima, we
additionally considered quadratic and quartic fits, as well as the
physics-informed form $EPM=a/V_x+b+cV_x^2$, motivated by the standard
induced, profile, and parasite power decomposition 
~\cite{Leishmann2000, Zeng2019}. Across the five resolved cases, the resulting
$V_x^*$ values differ by only 2.7--4.9\%, indicating that the choice of
fitting function does not significantly affect the identified optima.

\section{Discussion of Results}

\subsection{Significance of Findings}
\label{Imp}

First, the linear scaling of the energy-optimal velocity $V^*$ with the square root of the 
total vehicle mass $m$ enables simple yet effective energy-efficient flight control protocol. 
Specifically, the energy-optimal velocity can be obtained as 
\begin{equation}
  V_x^* = \sqrt {{\frac{mg}{C_{BD}}\tan \Theta^*}},  
  \label{eq: v_x*}
\end{equation}
where the mass-invariant $\Theta^*$ only needs to be solved once, 
and then plugged in to obtain the $V_x^*$ under any mass $m$.
Alternatively, practitioners can manage to measure/identify $V_x^*$ under one specific $m$ through experiments \cite{wu2020inflight, di2015energy},
and then scale with $\sqrt{m}$ to obtain $V_x^*$ under other mass conveniently.
Consequently, the burden of repeated experimental measurements and/or online adaptation is significantly reduced. 

Second, for multicopter energy consumption prediction and energy storage sizing, the found linear scaling 
of optimal $EPM$ with mass $m$ represents a highly useful advancement in practice. 
Specifically, the minimum energy consumption $E^*$ over a certain distance $L$ is 
\begin{equation}
  E^* = EPM^*L = C m L, 
  \label{eq: OptE}
\end{equation}
which can be used to determine the minimum energy capacity needed for the specified mission distance. 
Conversely, given a fixed capacity of the onboard energy storage (e.g. battery) $E$, 
the maximum range capability $L^*$ is
\begin{equation}
  L^* = \frac{E}{Cm}. 
\end{equation}
The constant $C$ can be either computed theoretically based on Eqn. (\ref{eq: C}) or identified from experimental testing. 
These simple formulas enable quick yet accurate (theoretically proven) evaluation of the operation range or energy requirement.

Third, the closed-form energy model yields a lightweight cost metric for mission planning under payload variation, since each cruise segment has energy proportional to $(m_v+m)L$ when flown at the corresponding $V_x^\ast$. 
This can be used to inform payload allocation and route selection without repeatedly solving an inner-loop velocity optimization:
\begin{equation}
 \min_{\{(m_i, L_i)\}_{i=1}^{N_m}} E = \sum_i C(m_v+m_i)L_i,
\end{equation}
where $m_v$ is the vehicle mass. 
Again, it needs to be emphasized that all these advantages hinge on the condition that the multirotor is operated under the energy-optimal velocity which varies with the mass, but not any arbitrary velocity. 

Finally, we remark that several commonly used mission-energy models in current literature directly assume a proportional relationship between vehicle mass and energy consumption rate $EPM$ under any horizontal velocity in the form of
\begin{equation}
EPM \propto f(V_x)m^a, 
\label{EPM_ref}
\end{equation} 
e.g. $a = 1$ in \cite{FIGLIOZZI2017251} and $a = 1.5$ 
in \cite{Dorling201770}, 
and  $f(V_x)$ is a certain function of the velocity.
Our results instead show that a simple linear-in-mass scaling holds only at the energy-optimal velocity, 
while the full $EPM$--$V_x$ curve does not scale uniformly with $m$. 
The invariant flight state governing normalized efficiency is the pitch angle at optimum, $\Theta^\ast$.

\subsection{Impacts of Other Component Efficiency}
\label{OCE}

As mentioned previously, other multicopter components, including the electric motor and the electronic speed controller (ESC), are not explicitly considered in the theoretical derivations of this work. Their efficiencies are treated as approximately constant over the flight regime of interest and therefore act as near-uniform multiplicative factors in the energy model, without altering the derived mass-scaling relationships.
This assumption is supported by experimental results in the literature \cite{michel2022performance}, which show that combined motor–ESC efficiency remains largely constant in the vicinity of the energy-optimal cruise velocity for a comparable multicopter platform, whereas the rotor aerodynamic efficiency (thrust per unit power) dominates the variation of the energy-per-meter (EPM) curve with respect to forward velocity.

\subsection{Scope and Limitations}
To clarify the scope of applicability of the derived scaling laws, we reiterate the key
assumptions underlying the analysis. The scaling laws hold for mass-only variation of a fixed
vehicle configuration in steady horizontal flight. Payload-induced center-of-mass shifts are
assumed negligible, preserving the equal rotor thrust distribution assumed in Sec. II-B. Changes
in airframe geometry or aerodynamic characteristics, e.g., the body drag coefficient, are also
not considered in the present scope. Furthermore, transient maneuvers involving acceleration
or attitude changes are excluded from the steady-state analysis.

\section{Conclusion}

This paper studies the scaling relationship between the energy-optimal flight behavior and multicopter mass
under the steady-state horizontal flight.
Several interesting and important findings have been made with both significant theoretical and practical implications.
First, the energy-optimal pitch angle is found to be invariant to mass.
This finding reflects the underlying mass-similarity property of steady horizontal flight: 
mass changes the scale of the steady-state solution, 
while the normalized aerodynamic state and the energy-optimal pitch angle remain unchanged.
Second, the energy-optimal velocity is found to be proportional to the square root of the total mass.
This finding enables convenient adaptation of flight control to mass/payload variation by
only requiring a simple scaling of the velocity. 
Third, the optimal energy efficiency, measured by energy consumption per distance traveled,
is found to be proportional to mass.
This finding significantly simplifies the mission/task level energy-optimal planning problem, e.g. payload assignment for drone delivery,
by enabling a simple linear problem formulation.
The results not only advance the understanding of the fundamental multicopter energy dynamics, 
but also lay the foundation for energy-efficient operation and control optimization.

\bibliographystyle{IEEEtran}
\bibliography{references}

\vfill

\end{document}